\documentclass[letterpaper, 10 pt, conference]{ieeeconf}

\IEEEoverridecommandlockouts
\usepackage{booktabs}
\usepackage{graphicx}
\usepackage{amsmath}
\usepackage{amssymb}
\usepackage{array}
\usepackage{url}
\usepackage[table,dvipsnames]{xcolor}
\usepackage{makecell}
\usepackage{pifont}
\usepackage{tabularx}
\usepackage{multirow}
\usepackage{capt-of}
\usepackage[colorlinks=true,urlcolor=blue,linkcolor=black,citecolor=black,breaklinks=true]{hyperref}

\renewcommand{\footnoterule}{\kern-3pt\hrule width 0.4\columnwidth\relax\kern 2.6pt}

\newcommand{\methodname}{AECNav: Active Evidence Consolidation for Efficient Zero-Shot Open-Vocabulary Object Navigation}
\newcommand{\cmark}{\ding{51}}
\newcommand{\xmark}{\ding{55}}
\newcommand{\NA}{--}
\newcommand{\best}[1]{\textbf{#1}}
\newcommand{\second}[1]{\underline{#1}}
\newlength{\figonepanelwd}
\newlength{\figonepanelht}
\newcommand{\figoneplaceholder}[1]{%
  \begingroup
  \setlength{\fboxsep}{0pt}%
  \fbox{%
    \colorbox{gray!8}{%
      \parbox[c][\figonepanelht][c]{\figonepanelwd}{%
        \centering\scriptsize\textcolor{gray!55}{#1}%
      }%
    }%
  }%
  \endgroup
}
\newcommand{\figonepanel}[2]{%
  \begingroup
  \setlength{\fboxsep}{0pt}%
  \IfFileExists{#1}{%
    \fbox{%
      \parbox[c][\figonepanelht][c]{\figonepanelwd}{%
        \centering
        \includegraphics[width=\figonepanelwd,height=\figonepanelht,keepaspectratio]{#1}%
      }%
    }%
  }{%
    \figoneplaceholder{#2}%
  }%
  \endgroup
}

\title{\LARGE \bf
\methodname
}

\author{%
\authorblockN{Guanlin Liu$^{1}$,
Shaobin Ling$^{1}$,
Renyuan Liu$^{1}$,
Zeying Gong$^{2}$,
Junjie Hu$^{1,\dagger}$}
\authorblockA{$^{1}$The Chinese University of Hong Kong, Shenzhen\\
$^{2}$The Hong Kong University of Science and Technology (Guangzhou)%
\thanks{\textbf{$^{\dagger}$Corresponding author.}}%
\thanks{E-mails: guanlinliu@link.cuhk.edu.cn, $^{\dagger}$hujunjie@cuhk.edu.cn}%
}%
}

\IEEEaftertitletext{%
  \begin{minipage}{\textwidth}
    \centering
    \includegraphics[
      width=1.0\linewidth
    ]{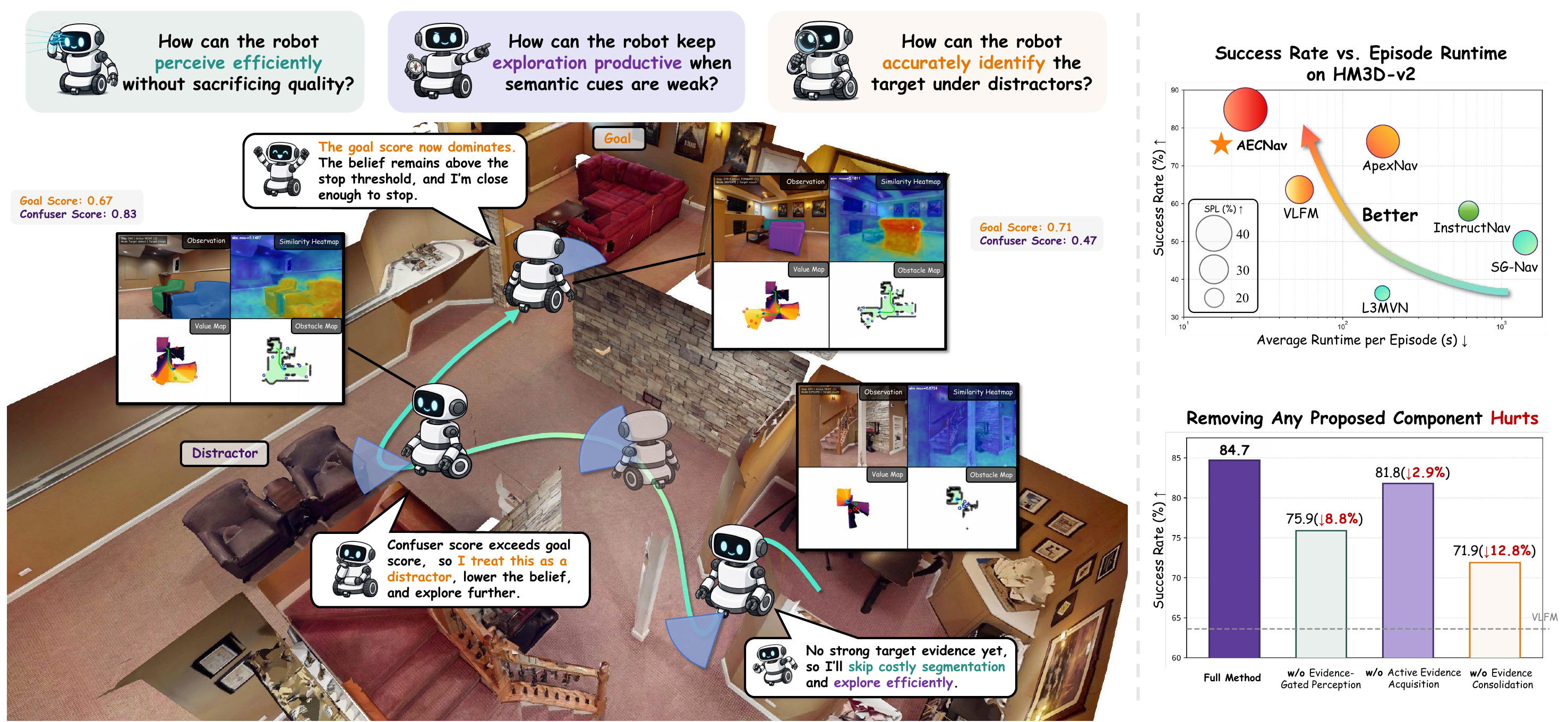}
       \vspace{-5mm}
    \captionof{figure}{\textbf{Overview of AECNav.} \textbf{Left:} the three questions an evidence-driven navigator must answer, and how AECNav answers them by integrating efficient perception, productive exploration, and reliable target confirmation into a navigation loop. \textbf{Top right:} AECNav attains a strong success--runtime trade-off. \textbf{Bottom right:} each of the three components answers one of these questions; removing any of them costs $2.9$ to $12.8$ points of success rate.
}
    \label{fig:intro}
    \vspace{0.2em}
  \end{minipage}%
}

\begin{document}

\bstctlcite{BSTctl}

\maketitle
\thispagestyle{empty}
\pagestyle{empty}

\begin{abstract}
% \textcolor{ForestGreen}{
Zero-shot object-goal navigation (ZSON) in open-vocabulary scenarios is challenging, as it requires a robot to locate an arbitrarily specified object in an unseen environment without task-specific training. Currently, the task still suffers from high latency and limited accuracy due to redundant perception pipelines and insufficient evidence for reliable target confirmation. In this letter, we reframe ZSON as an evidence-driven perception-to-decision problem and present AECNav, a training-free pipeline built on three components: i) Evidence-gated perception, which utilizes a shared encoding across all reasoning stages to establish a unified semantic basis and eliminate redundant computations; ii) Evidence consolidation, which aggregates detections into cluster-level log-odds beliefs. This explicitly separates genuine target support from the false confidence of visually similar distractors, while treating the absence of expected detections as negative evidence; and iii) Active evidence acquisition, which sustains productive exploration under weak semantic cues by selecting frontiers that maximize information gain at minimal traversal cost. 
As a result, AECNav significantly outperforms previous methods and achieves state-of-the-art success rates of 84.7\%, 57.3\%, and 51.3\% on HM3D-v2, HM3D-OVON, and MP3D, respectively, with substantially lower inference overhead, and attains 95\% success across 40 trials on a physical quadruped robot at roughly 5\,Hz. 
% Code will be made publicly available upon acceptance.  
\textbf{Project page:} \url{https://basaermi.github.io/aecnav-website/}.
\end{abstract}

\section{Introduction}

Zero-shot object-goal navigation (ZSON) requires a mobile robot to locate a specified object in a previously unseen environment, without task-specific training on either the target object categories or the scene layouts~\cite{gadre2023cows}. This challenge becomes even more pronounced in practical open-vocabulary settings, where the robot must ground arbitrary language goals without relying on category-specific detectors, while also rejecting visually similar distractors~\cite{yokoyama2024hm3dovon}.

Effective and practical ZSON requires a robot to accurately and efficiently localize the target object while selecting informative exploration actions at each time step. However, prior methods commonly face several key bottlenecks.
First, existing approaches often employ separate encoding pipelines for frontier selection and target-object confirmation~\cite{yokoyama2023vlfm,zhang2025apexnav}.  This design prevents different stages of visual reasoning from sharing a unified semantic foundation and forces each observation to be redundantly processed by multiple models at every navigation step, resulting in substantial inference latency.  Second, in open-vocabulary target confirmation, a high confidence score is not inherently self-explanatory: it may indicate genuine target evidence, semantic ambiguity from a visually similar object, or a transient false detection. 
However, reliably distinguishing the true target from similar distractors remains challenging.
Moreover, exploration is often guided by semantic relevance or geometric frontier utility, with limited explicit modeling of how much new evidence a reachable frontier is expected to provide relative to its traversal cost~\cite{gadre2023cows,yokoyama2023vlfm,zhang2025apexnav}. Together, these limitations hinder reliable target confirmation and efficient navigation under open-vocabulary uncertainty.

Motivated by these insights, we introduce the \textbf{A}ctive \textbf{E}vidence \textbf{C}onsolidation \textbf{Nav}igation (\textbf{AECNav}), a structured pipeline that integrates unified and efficient semantic perception, explicit spatial evidence accumulation, and active exploration for evidence-driven zero-shot ObjectNav.  AECNav consists of three core components:

i) Evidence-Gated Perception: To overcome the redundant semantic detections in prior works, we replace per-frame detection pipelines with a unified, on-demand perception architecture. 
Crucially, we introduce an evidence-gated triggering mechanism: the computationally expensive instance segmentation branch is kept dormant and dynamically invoked only when patch-level evidence identifies a highly probable target region. Overall, this module improves target confirmation accuracy while drastically reducing redundant visual processing overhead.

ii) Evidence Consolidation: To robustly confirm open-vocabulary targets, instance detections are seamlessly consolidated into 3D spatial clusters using an additive log-odds formulation. Unlike naive score-averaging, this update mechanism naturally separates genuine target support from the false confidence induced by visually similar distractors. Crucially, it also treats missing expected detections as negative evidence to suppress unsupported target hypotheses.
   
iii) Active Evidence Acquisition: When the accumulated evidence remains insufficient for a reliable decision, AECNav treats exploration as active evidence gathering. It selects reachable frontiers by jointly considering semantic relevance, expected spatial information gain along traversable paths, and traversal cost, keeping exploration productive even when semantic cues are weak. This strategy directs the robot toward viewpoints most likely to reduce uncertainty and expose decisive target evidence.

To fairly evaluate our method, we provide experimental results on benchmark datasets and
real-world scenarios.
As shown in Fig.~\ref{fig:intro},
AECNav significantly outperforms the previous state-of-the-art while incurring lower inference overhead.
Real-world robot experiments further demonstrate the practical effectiveness of  AECNav in physical environments.

\begin{figure*}[t]
    \centering
    \includegraphics[width=1.0\textwidth]{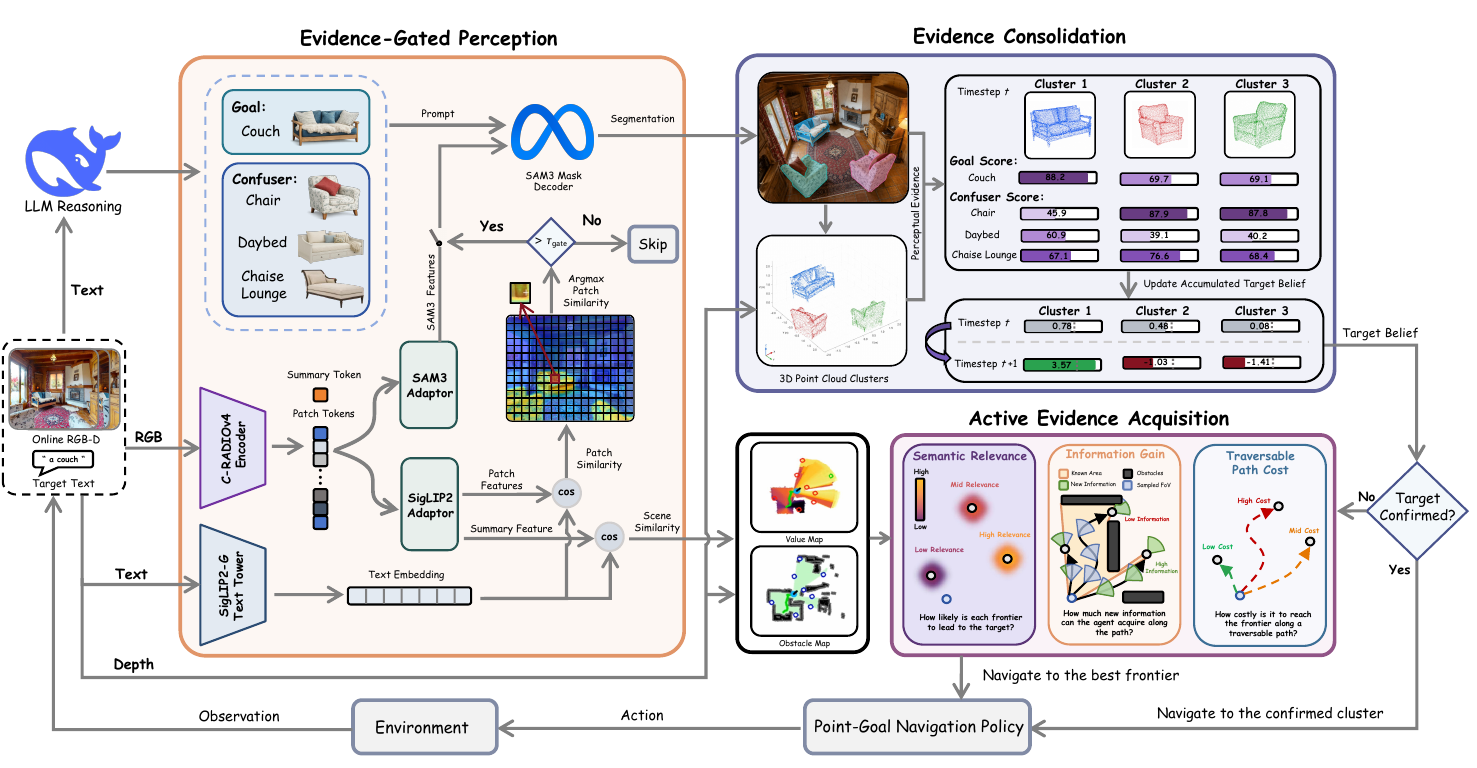}
        \vspace{-8mm}
    \caption{
    \textbf{Framework of AECNav.} AECNav comprises three modules: evidence-gated perception, which uses a shared C-RADIOv4 encoder to score scene relevance and trigger SAM3 only when a probable target is observed; evidence consolidation, which back-projects detections into 3D clusters and updates their log-odds beliefs using goal, confuser, and miss evidence; and active evidence acquisition, which selects frontiers by jointly considering semantic relevance, information gain, and traversal cost until a target cluster is confirmed. Confuser categories are generated offline by an LLM.}
    \label{fig:system-overview}
\end{figure*}

\section{Related Work}
 
Early methods such as \cite{majumdar2022zson} still require pretraining on the image navigation task, whereas CoW~\cite{gadre2023cows} removes this requirement by combining CLIP-based~\cite{clip} open-vocabulary localization with classical frontier exploration~\cite{yamauchi1997frontier}. This formulation has since become a common backbone for ZSON.
VLFM~\cite{yokoyama2023vlfm}, as a representative example, ranks frontiers using a semantic value map derived from image-text similarity to the target category. 
Our method also follows this paradigm; below, we present the key differences between our method and previous methods regarding perception pipeline design, target confirmation strategy, and frontier selection.

\subsection{Perception Pipelines for Navigation} 
 Existing value-map-based open-vocabulary ZSON approaches~\cite{yokoyama2023vlfm,zhang2025apexnav,gong2026stairway} typically employ disjoint visual models for exploration (e.g., CLIP~\cite{clip}, BLIP-2~\cite{blip2}) and target confirmation (e.g., GroundingDINO~\cite{liu2024grounding}, MobileSAM~\cite{mobile_sam}), resulting in redundant encoding and fragmented semantics.
 To eliminate this, we use a single C-RADIOv4 encoder~\cite{ranzinger2026cradiov4} to jointly extract scene-level context for frontier mapping and patch-level features for target confirmation. While recent methods~\cite{alama2025rayfronts,r2f2026} also use RADIO~\cite{ranzinger2024amradio} as a perception backbone, they rely solely on patch-level features for frontier scoring, offering limited guidance for out-of-view targets. To further reduce latency, we introduce an evidence-gated perception strategy that dynamically invokes the perception decoder only for highly probable targets. 
 
\subsection{Target Confirmation}

Target confirmation in ZSON has advanced from single-frame thresholds to temporal reasoning. Early systems stop when a detection score exceeds a fixed threshold~\cite{yokoyama2023vlfm}. Later methods strengthen confirmation with scene-graph credibility and re-perception in SG-Nav~\cite{yin2024sgnav}, VLM-based rejection in TriHelper~\cite{zhang2024trihelper}, and occlusion-free viewpoint selection in MSGNav~\cite{huang2026msgnav}. However, they still make one-shot candidate decisions, without accumulating evidence over time or explicitly handling visually similar distractors.
ApexNav~\cite{zhang2025apexnav} addresses this by fusing target-centric observations over time and storing visually similar objects. Its fusion, however, averages evidence rather than accumulating it: confidence cannot grow beyond the strongest single-frame score even when multiple observations agree, and observations are weighted by visible extent rather than informativeness, so ambiguous views can contribute as much as decisive ones.

In this letter, we introduce a log-odds belief model inspired by occupancy-grid mapping~\cite{elfes1989using, thrun2005probabilistic}, but shift the latent variable from cell occupancy to cluster identity. The model maintains, for each 3D candidate, a cumulative belief that it is the queried target. Each observation contributes an additive, decisiveness-weighted update, decomposed into target support, distractor-induced negative evidence, or corrective miss evidence. This converts target confirmation from thresholded recognition into evidence accumulation over object hypotheses, enabling confidence to grow with repeated consistent observations while explicitly suppressing visually similar distractors.

\subsection{Frontier Selection}
Frontier selection in ZSON has evolved from geometric heuristics to semantic guidance. Classical frontier exploration targets the nearest boundary of explored space, an approach retained by CoW~\cite{gadre2023cows}. VLFM~\cite{yokoyama2023vlfm} instead ranks frontiers using a vision-language value map. However, semantic ranking primarily indicates which direction is likely to contain the goal; it does not account for the amount of unknown space a frontier may reveal or the cost of reaching it. As a result, under weak or ambiguous semantic cues, agents may oscillate among distant, low-yield frontiers. ApexNav~\cite{zhang2025apexnav} addresses this issue by reverting to geometry-based exploration when semantic cues are unreliable. ASCENT~\cite{gong2026stairway} integrates both signals, ranking frontiers by semantic value and traversal cost, and invoking an LLM when fine-grained selection is needed. A related probabilistic line, including IPPON~\cite{qu2024ipponcommonsenseguided} and BeliefMapNav~\cite{zhou2026beliefmapnav}, selects paths or frontier sequences that maximize the expected probability of observing the target. Our method instead scores each frontier by jointly considering semantic value, traversal cost, and en-route information gain, enabling efficient exploration and reducing oscillation under weak semantic cues.

\section{Methodology}

We address the open-vocabulary ZSON task, where a robot must navigate to a target object specified by an arbitrary category name $g$ in previously unseen environments. At each time step $t$, the robot receives an egocentric RGB-D observation $(I_t, D_t)$ and its pose $T_t$ relative to the starting position. The episode is considered successful if the robot issues a stop action within a predefined distance of the target object before exceeding a fixed step budget. 
Fig.~\ref{fig:system-overview} illustrates the overall pipeline of AECNav. 

\subsection{Evidence-Gated Perception} \label{sec:perception}

AECNav extracts all visual cues from a single C-RADIOv4 encoder $E$ in one forward pass, producing a summary token and patch tokens: 
\begin{equation} 
\bigl(z_t^{\mathrm{sum}}, z_t^{\mathrm{patch}}\bigr) = E(I_t), 
\end{equation} 
where $z_t^{\mathrm{sum}}\in \mathbb{R}^{D}$ captures scene-level semantics and $z_t^{\mathrm{patch}} \in \mathbb{R}^{N \times D}$ encodes local visual content across $N$ spatial positions. The goal category $g$ is encoded once into a text embedding $e_g$ by the SigLIP2 text encoder~\cite{tschannen2025siglip2} and cached for the entire episode.

The summary token provides a global assessment of how relevant the current view is to the target. After the SigLIP2 adaptor $f$ aligns it with the text embedding space, its similarity to the goal embedding,
\begin{equation} 
\sigma_t^{\mathrm{scene}} = \cos\bigl(f(z_t^{\mathrm{sum}}), e_g\bigr), 
\end{equation}
is projected via depth and pose onto a top-down value map $V_t \in \mathbb{R}^{H \times W}$ that accumulates scene-level relevance over the trajectory. This map serves as a persistent spatial prior for frontier-based exploration (Sec.~\ref{sec:exploration}).

The summary token indicates whether the current scene is relevant but cannot localize where the target signal originates. The patch tokens resolve this. Applying the same adaptor $f$ to each patch feature yields per-patch similarities with the goal embedding, whose maximum serves as a localized target indicator:
\begin{equation} 
\sigma_t^{\mathrm{patch}} = \max_{1 \le i \le N}\cos\bigl(f(z_{t,i}^{\mathrm{patch}}), e_g\bigr). 
\end{equation} 
Since the patch-level similarity $\sigma_t^{\mathrm{patch}}$ already provides a strong preliminary signal for target presence, we introduce an evidence-gated mechanism to further optimize perceptual efficiency. Specifically, when
$\sigma_t^{\mathrm{patch}}$ exceeds a similarity threshold $\tau_{\mathrm{gate}}$, AECNav routes the shared visual features into the SAM3 decoder without re-encoding the image, producing instance-level detections for spatial evidence consolidation (Sec.~\ref{sec:belief}); otherwise, the segmentation step is skipped entirely. Since the patch similarities are already computed, this gating decision is nearly free, and in practice it skips over 60\% of SAM3 calls, reducing per-step latency by roughly 30\% (see Sec.~\ref{sec:efficiency} for detailed profiling).

\subsection{Evidence Consolidation}
\label{sec:belief}

When the segmentation gate triggers SAM3, each detected instance is back-projected via depth into a 3D point cloud and matched to the nearest recently observed cluster based on geometric proximity. If the geometric distance to this cluster falls within a fixed association radius, the instance's point cloud is fused into it; otherwise, the instance initializes a new cluster. Each cluster $\mathcal{C}_k$ maintains a scalar log-odds belief over the target hypothesis:
\begin{equation}
  l(\mathcal{C}_k) = \log \frac{P(\mathcal{C}_k = g)}{1 - P(\mathcal{C}_k = g)},
\end{equation}
where $k$ indexes the clusters and  $P(\mathcal{C}_k = g)$ denotes the estimated posterior probability that cluster $\mathcal{C}_k$ is an instance of the goal category $g$; positive values favor the target and negative values favor rejection. New clusters start at the neutral value $l(\mathcal{C}_k) = 0$ and are immediately refined by the semantic update from the same observation. Following the additive update structure of classical log-odds filtering~\cite{elfes1989using, thrun2005probabilistic}, the belief of each observed cluster is updated upon SAM3 invocation by accumulating semantic and miss evidence:
\begin{equation}
\label{eq:belief-update}
  l(\mathcal{C}_k) \leftarrow l(\mathcal{C}_k)
    + \Delta l^{\mathrm{sem}}(\mathcal{C}_k)
    + \Delta l^{\mathrm{miss}}(\mathcal{C}_k),
\end{equation}
with the accumulated value clamped to $[l_{\min},\,l_{\max}]$ to prevent saturation. The semantic term
$\Delta l^{\mathrm{sem}}$ encodes evidence from detected
instances, while the miss term $\Delta l^{\mathrm{miss}}$ encodes evidence
from the absence of expected detections.

To distinguish the target from visually similar distractors, an LLM generates a small set of confuser categories (up to three) for each target, and SAM3 is prompted with the goal and confuser categories simultaneously, so that each detection carries a goal score $s_g$ and the highest confuser score $s_{\mathrm{conf}}$.
Detections scoring below a fixed threshold \(\tau_{\mathrm{det}}\) are discarded and never recorded as instances. 
For the cluster matched by a detection, the semantic term in Eq.~\eqref{eq:belief-update} follows from comparing the two scores:
\begin{equation}
\label{eq:sem-update}
\Delta l^{\mathrm{sem}}(\mathcal{C}_k) =
\begin{cases}
+\alpha_{\mathrm{sem}}\,\rho_k\,\mathrm{logit}(\tilde{s}_g),
  & s_g \ge s_{\mathrm{conf}} + \delta, \\[4pt]
-\alpha_{\mathrm{sem}}\,\rho_k\,\mathrm{logit}(\tilde{s}_{\mathrm{conf}}),
  & s_{\mathrm{conf}} \ge s_g + \delta, \\[4pt]
0,
  & \text{otherwise},
\end{cases}
\end{equation}
where $\mathrm{logit}(x) = \log\frac{x}{1-x}$, 
\(\tilde{s}_g\) and \(\tilde{s}_{\mathrm{conf}}\) are rescaled versions of  $s_g$, $s_{\mathrm{conf}}$, ensuring that valuable detections (i.e., $s_g$, $s_{\mathrm{conf}}$ $\ge$ \(\tau_{\mathrm{det}}\)) yield non-negative logits.
$\rho_k \in [0{,}1]$ is a spatial weight given by the overlap between the new instance's point cloud and cluster $\mathcal{C}_k$, $\delta > 0$ defines a neutral margin, and $\alpha_{\mathrm{sem}}$ scales the strength of semantic evidence. When the goal score clearly dominates (first term in Eq.~\eqref{eq:sem-update}), the observation reinforces the target hypothesis; when the confuser score dominates (middle term), the same observation actively suppresses the cluster's belief. Ambiguous cases are left unchanged, preventing detection noise from corrupting the long-term estimate.

Not all informative signals come from detections. If a cluster with positive belief falls within the camera's field of view but is not detected in the current frame, this absence itself constitutes negative evidence. We define the visibility fraction
$v_k$ as the proportion of the cluster's 3D points that project into the current view frustum, yielding the miss term update for such clusters:
\begin{equation}
\label{eq:miss-update}
\Delta l^{\mathrm{miss}}(\mathcal{C}_k)
  = -\alpha_{\mathrm{miss}}\,v_k\,\mathrm{logit}(1 - s_{\mathrm{miss}}),
\end{equation}
where $\alpha_{\mathrm{miss}}$ is a weighting coefficient and $s_{\mathrm{miss}}$ is an empirical prior for the per-frame miss rate.  Clusters that should have been observed but were not are gradually corrected, while those outside the field of view remain unaffected.

The accumulated belief drives the stopping decision. 
Once a cluster's belief exceeds \(\tau_{\mathrm{stop}}\), AECNav navigates toward the cluster while continuously updating its belief from subsequent observations. The robot stops only when the belief exceeds \(\tau_{\mathrm{stop}}\) consistently over a temporal window and it is sufficiently close to the cluster. This requires both evidential stability and spatial proximity, preventing premature commitment from transient high-confidence detections.

\newcolumntype{L}[1]{>{\raggedright\arraybackslash}p{#1}}
\newcolumntype{C}[1]{>{\centering\arraybackslash}p{#1}}
\newcolumntype{Y}{>{\centering\arraybackslash}X}
\begin{table*}[t]
\centering
\vspace{3mm}
\caption{Comparison with previous methods on HM3D-v2, MP3D, and HM3D-OVON.}
\vspace{-3mm}
\label{tab:main-results}
\footnotesize

\begin{tabularx}{\textwidth}{
@{}
L{0.16\textwidth}
C{0.115\textwidth}
L{0.120\textwidth}
Y Y Y Y Y Y
@{}}
\toprule
\multirow{2}{*}{\textbf{Method}} &
\multirow{2}{*}{\textbf{Training-free}} &
\multirow{2}{*}{\textbf{Venue}} &
\multicolumn{2}{c}{\textbf{HM3D-v2}} &
\multicolumn{2}{c}{\textbf{MP3D}} &
\multicolumn{2}{c}{\textbf{HM3D-OVON}} \\
\cmidrule(lr){4-5}\cmidrule(lr){6-7}\cmidrule(l){8-9}
& & &
\textbf{SR$\uparrow$} & \textbf{SPL$\uparrow$} &
\textbf{SR$\uparrow$} & \textbf{SPL$\uparrow$} &
\textbf{SR$\uparrow$} & \textbf{SPL$\uparrow$} \\
\midrule

Uni-NaVid~\cite{zhang2025uninavid}& \xmark & RSS 2025 & 73.7 & 37.1 & \NA & \NA & 39.5 & 19.8 \\
MTU3D~\cite{zhu2025mtu3d} & \xmark & ICCV 2025 & \NA & \NA & \NA & \NA & 40.8 & 12.1 \\
CompassNav~\cite{Li_2026_CompassNav} & \xmark & ICLR 2026 & 61.6 & 27.8 & 42.0 & 17.5 & 43.5 & 21.6 \\
OVSegDT~\cite{zemskova2026ovsegdt} & \xmark & CVPR 2026 & \NA & \NA & \NA & \NA & 44.7 & 20.6 \\
TrajRAG~\cite{wang2026trajrag} & \xmark & CVPR 2026 & \second{78.1} & \second{40.2} & 42.6 & \second{18.0} & \NA & \NA \\

\midrule

CoW~\cite{gadre2023cows} & \cmark & CVPR 2023 & \NA & \NA & 9.2 & 4.9 & \NA & \NA \\
BeliefMapNav~\cite{zhou2026beliefmapnav} & \cmark & NeurIPS 2025 & \NA & \NA & 37.3 & 17.6 & \NA & \NA \\
ASCENT~\cite{gong2026stairway} & \cmark & IEEE RA-L 2026 & \NA & \NA & 44.5 & 15.5 & \NA & \NA \\
L3MVN~\cite{yu2023l3mvn} & \cmark & IROS 2023 & 36.3 & 15.7 & 34.9 & 14.5 & \NA & \NA \\
SG-Nav~\cite{yin2024sgnav} & \cmark & NeurIPS 2024 & 49.6 & 25.5 & 40.2 & 16.0 & \NA & \NA \\
InstructNav~\cite{long2025instructnav} & \cmark & CoRL 2024 & 58.0 & 20.9 & \NA & \NA & \NA & \NA \\
VLFM~\cite{yokoyama2023vlfm} & \cmark & ICRA 2024 & 63.6 & 32.5 & 36.4 & 17.5 & 35.2 & 19.6 \\
WMNav~\cite{Nie_2025_WMNav} & \cmark & IROS 2025 & 72.2 & 33.3 & \second{45.4} & 17.2 & \NA & \NA \\
MSGNav~\cite{huang2026msgnav} & \cmark & CVPR 2026 & 74.1 & 33.4 & \NA & \NA & \second{48.3} & \second{27.0} \\
ApexNav~\cite{zhang2025apexnav} & \cmark & IEEE RA-L 2025 & 76.2 & 38.0 & 39.2 & 17.8 & \NA & \NA \\

\midrule
\rowcolor{gray!15}
AECNav (ours) & \cmark & Ours &
\best{84.7} & \best{45.3} &
\best{51.3} & \best{25.9} &
\best{57.3} & \best{30.5} \\
\bottomrule
\end{tabularx}
\end{table*}

\subsection{Active Evidence Acquisition}
\label{sec:exploration}

When no cluster's belief is sufficient for commitment, the robot must decide where to gather new evidence. Semantic relevance indicates which direction may contain the target; especially when such cues are weak, exploration must gain the most new information at the least traversal cost, so that fresh semantic evidence can emerge to support the decision. AECNav therefore selects the frontier that maximizes a composite utility:
\begin{equation}
  U_t(f;\,g)
    = \widetilde{S}_t(f;\,g)
    + \lambda_{\mathrm{info}}\,\widetilde{G}_t(f)
    - \lambda_{\mathrm{dist}}\,\widetilde{C}_t(f),
        \label{eq_frontier}
\end{equation}
where $\lambda_{\mathrm{info}}$ and $\lambda_{\mathrm{dist}}$ are weighting coefficients, and the tilde $\widetilde{(\cdot)}$ denotes min-max normalization across the current frontier set, making terms of different scales directly comparable.
The semantic term $S_t$ aggregates goal-relevant value from the map around each frontier, as in VLFM~\cite{yokoyama2023vlfm}, pulling the robot toward regions likely to contain the target. The
information gain term $G_t$ estimates how much unobserved space the robot would reveal by traveling to the frontier along a feasible path. The cost term $C_t$ penalizes long traversals, discouraging expensive detours for marginal return.

For each candidate frontier $f$, AECNav computes the shortest traversable path $\pi_t(f)$ from the robot via BFS on the occupancy grid. Viewpoints are sampled at regular intervals along this path, with headings determined by the local path tangent; the endpoint heading is oriented toward nearby unknown cells. At each sampled viewpoint, rays are cast within the camera's horizontal field of view, traversing free and unknown cells and terminating upon hitting a known obstacle or reaching the sensing range; only unknown cells encountered along each ray contribute to the information gain.
%Fig.~\ref{fig:bfs-ray-oeg} illustrates this process. 
The information gain and traversal cost are then computed over the same path:
\begin{equation}
  G_t(f)
    = \bigl|\,\mathcal{U}_t \cap \mathrm{Vis}\bigl(\pi_t(f)\bigr)\,\bigr|,
  \qquad
  C_t(f)
    = \bigl|\pi_t(f)\bigr|.
\end{equation}
where $\mathcal{U}_t$ is the set of currently unknown cells, $\mathrm{Vis}(\pi_t(f))$ is the union of cells visible from all sampled viewpoints via ray casting, and $|\pi_t(f)|$ is the path length. This formulation estimates not only what the robot would observe at the destination, but also what it would see along the
way. The resulting observations re-enter the perception and belief stages, closing the loop: the robot explores when target belief is insufficient, and stops once accumulated evidence meets the stability criterion defined in Sec.~\ref{sec:belief}.

\section{Experiments}

\subsection{Experimental Settings}
\textbf{Datasets.}
For a fair comparison, we evaluate our approach against prior methods on three benchmarks in the Habitat simulator~\cite{savva2019habitat}: HM3D-v2~\cite{yadav2023hm3dsem}, comprising 1,000 test episodes, 36 scenes, and 6 goal categories; MP3D~\cite{chang2017matterport3d}, comprising 2,195 test episodes, 11 scenes, and 21 goal categories; and HM3D-OVON~\cite{yokoyama2024hm3dovon}, comprising 3,000 test episodes, 36 scenes, and 49 goal categories. Notably, HM3D-OVON is specifically designed for the open-vocabulary setting.

\textbf{Evaluation Metrics.}
We report two standard ObjectNav metrics: Success Rate (SR) and Success weighted by Path Length (SPL). All results follow the same settings as prior work~\cite{Nie_2025_WMNav,Li_2026_CompassNav,huang2026msgnav}, with episodes capped at 500 steps. For efficiency analysis, we additionally report per-episode steps and runtime, along with per-step latency in ablations. 

\textbf{Implementation Details.}
The Habitat simulator provides $640{\times}480$ RGB-D observations with a $79^{\circ}$ horizontal FoV and a depth range of $[0.5,\,5.0]$\,m. The camera is mounted at $0.88$\,m, and the robot selects from \textsc{move\_forward} ($0.25$\,m), \textsc{turn\_left/right} ($30^{\circ}$), and \textsc{stop}. 
Following VLFM~\cite{yokoyama2023vlfm}, we use a pretrained PointNav policy for low-level waypoint navigation.
The top-down occupancy and value maps share an $H{\times}W{=}1000{\times}1000$ grid at $0.05$\,m/cell. The shared backbone is C-RADIOv4 (SO400M) at $672{\times}672$ input resolution, with SigLIP2-G and SAM3 adaptor heads.
% We use C-RADIO-v4 (SO400M) at $672{\times}672$ resolution as the shared visual backbone, with SigLIP2-G and SAM3 adaptors providing text-aligned projections and instance segmentation respectively. 
The segmentation gate fires when the maximum patch similarity exceeds $\tau_{\mathrm{gate}}{=}0.08$, and SAM3 detections scoring below $\tau_{\mathrm{det}}{=}0.4$ are discarded. An instance is fused into an existing cluster when their geometric distance falls below $0.75$\,m. Confuser categories are generated by DeepSeek-V4-Flash and cached across episodes. Belief is bounded to $[l_{\min},l_{\max}]{=}[-4,4]$ with neutral margin $\delta{=}0.10$, evidence weights $(\alpha_{\mathrm{sem}}, \alpha_{\mathrm{miss}}){=}(0.7,0.3)$, and miss prior $s_{\mathrm{miss}}{=}0.2083$, estimated from detection statistics as the per-frame miss rate. The robot stops when $\tau_{\mathrm{stop}}{=}1.0$ is exceeded in ${\ge}2$ of $5$ consecutive frames and it is within $0.5$\,m of the target cluster.
Exploration weights are $(\lambda_{\mathrm{info}}, \lambda_{\mathrm{dist}}){=}(1.0,1.0)$, with rays cast at $5^{\circ}$ steps over $5.0$\,m range and viewpoints sampled at $0.75$\,m stride along the BFS path. All experiments run on a single NVIDIA RTX 4090 and Intel Core i9-14900K.

\subsection{Quantitative Results}

\textbf{Comparison on Navigation Accuracy.}
Table~\ref{tab:main-results} presents quantitative comparisons between our method and prior approaches. As shown, AECNav consistently outperforms all previous methods, achieving the highest SR and SPL across all three benchmarks without any task-specific training.

Specifically, on HM3D-v2, our method achieves 84.7\% SR and 45.3\% SPL, outperforming the previous state-of-the-art method TrajRAG~\cite{wang2026trajrag} by an absolute 6.6\% in SR and 5.1\% in SPL. On MP3D, AECNav reaches 51.3\% SR and 25.9\% SPL, exceeding WMNav~\cite{Nie_2025_WMNav} by 5.9\% in SR and 8.7\% in SPL. The large improvement in SPL suggests that our information-aware frontier selection effectively reduces unnecessary detours in the structurally complex MP3D scenes.
On HM3D-OVON, the most challenging open-vocabulary benchmark, our method achieves 57.3\% SR and 30.5\% SPL, surpassing MSGNav~\cite{huang2026msgnav} by 9.0\% in SR. This substantial improvement is consistent with the design of our log-odds belief mechanism, which explicitly handles ambiguity from visually similar distractors, a challenge that becomes increasingly prevalent as the goal vocabulary grows.

\textbf{Comparison on Efficiency.} \label{sec:efficiency}
Table~\ref{tab:episode-runtime} compares the average number of steps and end-to-end  runtime per episode, measured on the first 100 episodes of HM3D-v2. Methods that invoke an LLM online spend most of their time on language model reasoning: SG-Nav takes over 1,400\,s per episode and InstructNav over 600\,s, despite step counts comparable to other baselines. Among baseline methods without online LLM calls, VLFM attains the lowest prior runtime (53.36\,s), whereas ApexNav, executed through a ROS-based control loop, reaches 177.97\,s. AECNav completes an episode in 24.39\,s on average, $2.2\times$ faster than VLFM, while also taking the fewest steps (108.63, 33\% fewer than the second-best ASCENT). This gain draws on two independent sources: information-aware exploration shortens each episode, while the shared encoding and evidence-gated perception cut the cost of every step, as dissected below.

\begin{table}[t]
\centering
\vspace{3mm}
\caption{Efficiency comparison on the first 100 episodes of HM3D-v2}
\vspace{-3mm}
\label{tab:episode-runtime}
\begin{tabular}{llcc}
\toprule
\textbf{Method}
  & \textbf{LLM Effort}
  & \textbf{Steps }$\downarrow$
  & \textbf{Runtime (s) }$\downarrow$ \\
\midrule
SG-Nav\cite{yin2024sgnav}           & Per-step & 278.60 & 1407.53 \\
InstructNav\cite{long2025instructnav}                  & On-demand & 212.20 & 626.96  \\
L3MVN\cite{yu2023l3mvn}                        & On-demand & 187.90 & 177.54 \\
ASCENT\cite{gong2026stairway}       & On-demand & \second{162.09} & 200.31 \\
VLFM\cite{yokoyama2023vlfm}         & None & 169.43 & \second{53.36} \\
ApexNav\cite{zhang2025apexnav}      & Offline cached & 167.59 & 177.97 \\
\midrule
\rowcolor{gray!15}
AECNav                             & Offline cached & \best{108.63} & \best{24.39} \\
\bottomrule
\end{tabular}
\end{table}

\begin{table}[t]
\centering
% \vspace{-3mm}
\caption{Perception pipeline and gating analysis on HM3D-v2.
}
\vspace{-3mm}
\label{tab:gate-analysis}
\setlength{\tabcolsep}{4pt}
\begin{tabular}{lcccc}
\toprule
\textbf{Perception Setting}
  & \textbf{SR}$\uparrow$
  & \textbf{SPL}$\uparrow$
  & \makecell{\textbf{Latency}\\\textbf{(s/step)}$\downarrow$}
  & \textbf{Skip (\%)}\\
\midrule
BLIP-2 + YOLOv7 + MobileSAM
  & 75.9  & 36.6  & 0.402  & --  \\
C-RADIO (no gate)
  & \best{84.7}  & 45.2  & 0.248  & 0   \\
C-RADIO + gate ($\tau_{\mathrm{gate}}{=}0.12$)
  & 84.0  & 44.9  & \best{0.162}  & 78.5  \\
C-RADIO + gate ($\tau_{\mathrm{gate}}{=}0.10$)
  & 84.5  & 45.2  & 0.173  & 72.3  \\
\rowcolor{gray!15}
C-RADIO + gate ($\tau_{\mathrm{gate}}{=}0.08$)
  & \best{84.7}  & \best{45.3}  & 0.178  & 62.5  \\
\bottomrule
\end{tabular}
\end{table}

% \textbf{Perception and Gating.}
Table~\ref{tab:gate-analysis} breaks down the contributions of shared encoding and the segmentation gate. For a controlled comparison, all latencies in this analysis measure method inference alone, excluding simulator stepping and benchmarking overhead. Replacing the multi-model pipeline with C-RADIO raises SR by 8.8\% and SPL by 8.6\% at $1.6{\times}$ lower latency: a unified semantic basis lets all reasoning stages draw on the same
visual representation, eliminating inconsistencies introduced by separate encoding paths.
Adding a segmentation gate effectively skips unnecessary calls, presenting a clear accuracy--speed tradeoff: at
$\tau_{\mathrm{gate}}{=}0.12$, 78.5\% of calls are skipped yet SR drops by 0.7\%; as the threshold relaxes to
$\tau_{\mathrm{gate}}{=}0.08$, skip rate settles at 62.5\% and SR fully recovers to 84.7\%. We therefore adopt
$\tau_{\mathrm{gate}}{=}0.08$, which preserves full accuracy at $2.3{\times}$ the speed of the multi-model baseline.

\subsection{Ablation Studies}
\begin{table}[t]
\centering
\vspace{3mm}
\caption{Ablation study of the three core components on HM3D-v2.}
\vspace{-3mm}
\label{tab:ablation}
\setlength{\tabcolsep}{6pt}
\begin{tabular}{lcc}
\toprule
\textbf{Setting}
  & \textbf{SR}$\uparrow$
  & \textbf{SPL}$\uparrow$ \\
\midrule
w/o Evidence-Gated Perception
  & 75.9 & 36.6 \\
w/o Evidence Consolidation
  & 71.9 & 40.5 \\
w/o Active Evidence Acquisition
  & 81.8 & 41.8 \\
\midrule
\rowcolor{gray!15}
AECNav (full)
  & \textbf{84.7} & \textbf{45.3} \\
\bottomrule
\end{tabular}
\end{table}

\begin{table}[!t]
\centering
% \vspace{-3mm}
\caption{Ablation of negative evidence types in evidence consolidation on HM3D-v2.}
\vspace{-3mm}
\label{tab:ablation-evidence}
\setlength{\tabcolsep}{6pt}
\begin{tabular}{ccccc}
\toprule
\multirow{2}{*}{\textbf{Goal}}
  & \multicolumn{2}{c}{\textbf{Negative Evidence}}
  & \multirow{2}{*}{\textbf{SR}$\uparrow$}
  & \multirow{2}{*}{\textbf{SPL}$\uparrow$} \\
\cmidrule(lr){2-3}
 & \textbf{Confuser} & \textbf{Miss} & & \\
\midrule
\cmark & \xmark & \xmark &81.9 & 43.9 \\
\cmark & \cmark & \xmark & 83.7 & 44.5 \\
\cmark & \xmark & \cmark & 83.2 & 45.1 \\
\rowcolor{gray!15}
\cmark & \cmark & \cmark & \textbf{84.7} & \textbf{45.3} \\
\bottomrule
\end{tabular}
\end{table}

\textbf{Ablation of Main Components.}
Table~\ref{tab:ablation} highlights ablation experiments on the main components. Here ``w/o'' means the corresponding module falls back to a conventional alternative or is removed altogether.
\paragraph{w/o Evidence-Gated Perception} scene scoring, detection, and segmentation are handled by separate BLIP-2~\cite{blip2}, YOLOv7~\cite{wang2023yolov7} and MobileSAM~\cite{mobile_sam} encoders. This variant causes an 
8.8\% drop in SR and an 
8.7\% drop in SPL, showing that a shared semantic basis across reasoning stages is important for both accuracy and efficiency.
\paragraph{w/o Evidence Consolidation} the robot navigates to the nearest target point cloud as soon as one is detected, without evidence accumulation. It causes the largest drop ($-12.8\%$ SR), confirming that structured evidence accumulation is critical for accurate target confirmation; without it, the robot cannot reliably model target confidence under similar-object interference and transient false detections.
\paragraph{w/o Active Evidence Acquisition} frontier selection uses semantic relevance only ($\lambda_{\mathrm{info}}{=}\lambda_{\mathrm{dist}}{=}0$). It costs $2.9\%$ SR and $3.5\%$ SPL, suggesting that reasoning about observational value and reachability beyond semantic relevance improves both navigation success and efficiency.

\textbf{Ablation of Evidence Consolidation.}

Table~\ref{tab:ablation-evidence} details the ablation results evaluating the confuser (middle term in Eq.\eqref{eq:belief-update}) and the missing detection (Eq.\eqref{eq:miss-update}) within evidence consolidation. Relying solely on the log-odds belief with goal score (first term in Eq.\eqref{eq:belief-update}) yields an 81.9\% SR, a 10\% improvement over the 71.9\% SR observed without evidence consolidation.
Accumulation itself thus carries most of the benefit: consistent observations compound into a running belief, so a candidate must earn confirmation across viewpoints rather than be accepted on first detection. 
The remaining two terms add 
2.8\% to the SR by repairing distinct failures. The confuser term gains 
1.8\% by actively suppressing distractors (e.g., preventing the robot from mistaking a sofa for a chair). The missing detection term gains 
1.3\% by undoing the belief raised by false positives, releasing the robot back to exploration when a cluster stops reappearing. Because these gains are nearly additive, they clearly correct disjoint errors, together raising AECNav to an 84.7\% SR.

\begin{figure}[t]
  \centering
  \vspace{3mm}
  \includegraphics[width=1.0\linewidth]{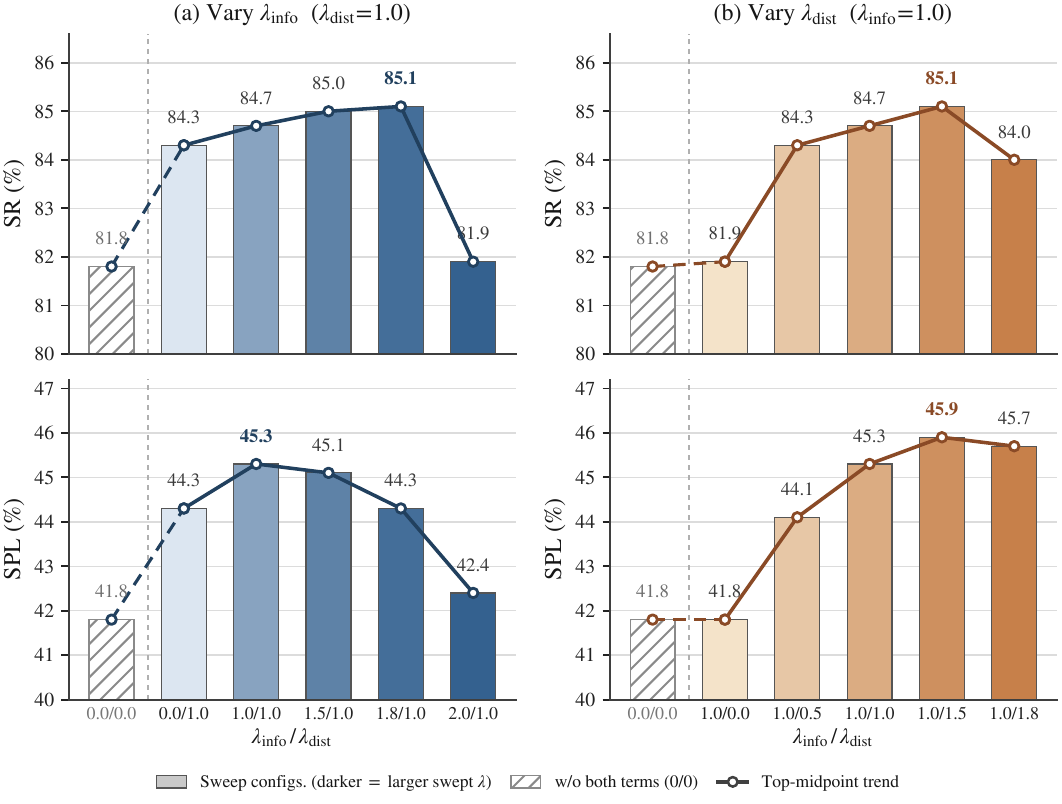}
  \vspace{-6mm}
  \caption{Results on varying the weights for information gain $\lambda_{\mathrm{info}}$ and traversal cost
  $\lambda_{\mathrm{dist}}$ in Eq.~\eqref{eq_frontier}. }
  \label{fig:ablation_lambda}
\end{figure}

\textbf{Ablation of Exploration Weights.}
We ablate the information gain $\lambda_{\mathrm{info}}$ and traversal cost $\lambda_{\mathrm{dist}}$ weights in Eq.~\eqref{eq_frontier}.
% by fixing one at its default value and varying the other. 
As shown in Fig.~\ref{fig:ablation_lambda}, enabling the traversal cost term alone already yields a marked improvement, lifting SR from 81.8\% to 84.3\% and SPL from 41.8\% to 44.3\%, and progressively adding information gain on top further raises SR to 85.1\%. In contrast, information gain alone brings virtually no benefit (81.9\% vs. 81.8\% SR); it becomes effective only when paired with the cost term. We attribute this to the nature of the information objective: without a notion of cost, the utility consistently favors large open regions with wide visibility, sending the robot on long excursions that consume the step budget while drifting away from semantically promising areas. Over-weighting either term is harmful: $\lambda_{\mathrm{info}}{=}2.0$ dilutes semantic guidance and collapses SR to 81.9\%, while an overly large $\lambda_{\mathrm{dist}}$ discourages necessary long-range exploration.
Although \(\lambda_{\mathrm{dist}}=1.5\) performs slightly better on HM3D-v2, the gain is not consistent across benchmarks; we therefore adopt   \(\lambda_{\mathrm{info}}=1.0\) and \(\lambda_{\mathrm{dist}}=1.0\) as the default setting across all datasets to avoid dataset-specific tuning.

\begin{figure*}[t]
    \centering
    \includegraphics[width=1.0\textwidth]{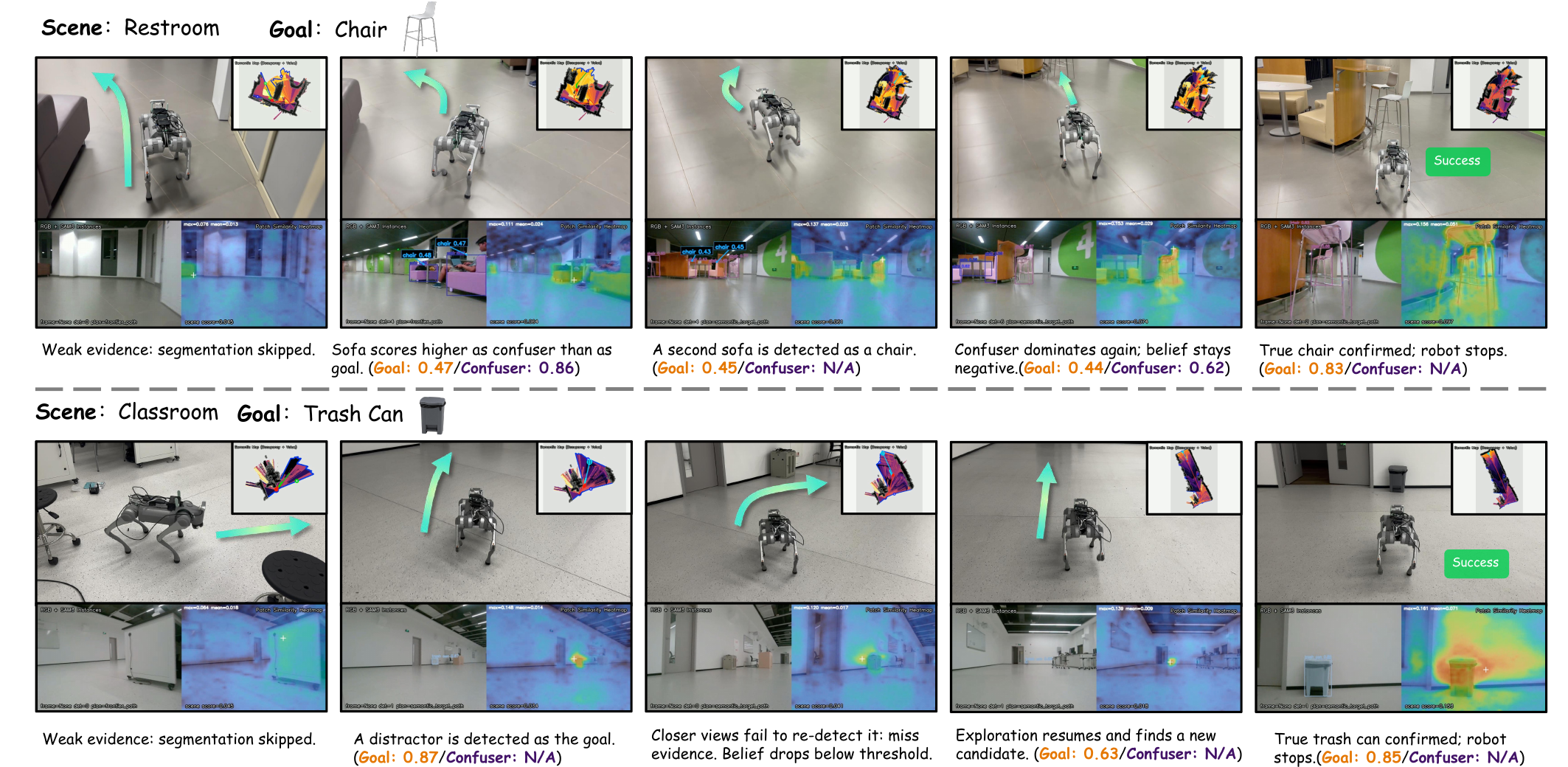}
        \vspace{-3mm}
    \caption{\textbf{Real-world trials on a Unitree Go2.} Top: confuser evidence rejects two sofas while searching for a chair. Bottom: miss evidence withdraws a distractor that fails to reappear at close range, and the robot then reaches the true object.}
    % We test AECNav in simulate and real-world environments.}
    \vspace{-5mm}
    \label{fig:real_world}
\end{figure*}

\subsection{Real-World Deployment}

We deploy AECNav on a Unitree Go2 quadruped robot equipped with an Intel RealSense D455 RGB-D camera. 
All perception and planning run on the same workstation, connected to the robot via binary WebSocket.
We evaluate eight open-vocabulary targets across four indoor scenes with five trials each, yielding 40 episodes in total; notably, several targets, such as water refill station and coffee machine, fall outside standard benchmark vocabularies. Moreover, the chair trials are deliberately staged with nearby sofas and benches, stress-testing the method's ability to distinguish the target from visually similar distractors.

Table~\ref{tab:real-world} reports the per-target success rate (SR) together with the average traveled distance (TD) and traveled time (TT) over successful episodes. AECNav succeeds in 38 of 40 episodes (95\%), achieving full success on six of the eight targets, with TD ranging from about 8\,m in the compact office to over 26\,m in the classroom scene. 
In the two failures, the coffee machine lay in a corner never observed by the forward-facing camera and the elevator trial exhausted the step budget in a large open area. 
As shown in Table~\ref{tab:real-latency}, AECNav sustains a decision rate of roughly 5\,Hz ($197$\,ms per decision) in real-world deployment.

Fig.~\ref{fig:real_world} illustrates the behavior of our method in two representative cases. Full demonstrations for all targets across the four scenes are provided in the supplementary video.

\begin{table}[t]
\centering
\vspace{3mm}
\caption{Real-world navigation results.}% on a Unitree Go2 robot. equipped with a RealSense D455 camera.}
\vspace{-3mm}
\label{tab:real-world}
\setlength{\tabcolsep}{5pt}
\begin{tabular}{llccc}
\toprule
\textbf{Scene} & \textbf{Target} & \textbf{SR} & \textbf{TD (m)} & \textbf{TT (s)} \\
\midrule
\multirow{2}{*}{Office}
  & Sofa            & 5/5 & 8.33  & 26.92 \\
  & Coffee machine  & 4/5 & 13.48 & 37.55 \\
\midrule
\multirow{2}{*}{Pathway}
  & Water refill station & 5/5 & 9.43  & 28.11 \\
  & Printer         & 5/5 & 19.90 & 44.13 \\
\midrule
\multirow{2}{*}{Restroom}
  & Chair           & 5/5 & 20.03 & 51.17 \\
  & Potted plant    & 5/5 & 21.57 & 49.87 \\
\midrule
\multirow{2}{*}{Classroom}
  & Elevator        & 4/5 & 23.31 & 53.44 \\
  & Trash can       & 5/5 & 26.14 & 51.96 \\
\midrule
\multicolumn{2}{l}{\textbf{Overall}} & 38/40 & 17.77 & 42.89 \\
\bottomrule
\end{tabular}
\end{table}

\begin{table}[t]
\centering
\footnotesize
% \vspace{-3mm}
\caption{Per-decision latency breakdown of the deployment pipeline, averaged over all real-world episodes.}
\vspace{-3mm}
\label{tab:real-latency}
\setlength{\tabcolsep}{3pt}
\begin{tabular}{lcccccc}
\toprule
 & \textbf{Sensing} & \textbf{Mapping} & \textbf{Percep.} & \textbf{Planning} & \textbf{Sched.} & \textbf{Total} \\
\midrule
\textbf{Latency (ms)} & 33.9 & 19.2 & 56.3 & 72.2 & 15.8 & \textbf{197.4} \\
\bottomrule
\end{tabular}
\end{table}

\section{Conclusion}

We present AECNav, which formulates ZSON as an evidence-driven perception-to-decision problem and addresses it through three tightly coupled components. First, evidence-gated perception derives all semantic cues from a shared encoding and invokes segmentation only when local evidence justifies it. Second, evidence consolidation accumulates detections into a running cluster-level belief, explicitly distinguishing positive, confuser, and miss evidence. Finally, when the accumulated evidence remains insufficient, active evidence acquisition guides exploration toward regions expected to provide the greatest new information at the lowest traversal cost. AECNav achieves state-of-the-art performance on three benchmark datasets with substantially improved efficiency, while real-world deployment at approximately 5\,Hz further demonstrates its practical effectiveness.

\bibliographystyle{IEEEtran}
\bibliography{IEEEabrv,references}

\end{document}